\documentclass{article}

\usepackage{microtype}
\usepackage{graphicx}
\usepackage{subfigure}
\usepackage{booktabs} 

\usepackage{dirtytalk}
\usepackage{subcaption}

\usepackage{hyperref}

\usepackage[accepted]{paper}

\usepackage{amsmath}
\usepackage{amssymb}
\usepackage{mathtools}
\usepackage{amsthm}

\usepackage{tabularx, array} 
\newcolumntype{Y}{>{\centering\arraybackslash}X}

\usepackage{xcolor}         
\definecolor{mBlue}{HTML}{4085CA}
\usepackage[most]{tcolorbox}
\usepackage[export]{adjustbox}

\usepackage[capitalize,noabbrev]{cleveref}

\theoremstyle{plain}

\theoremstyle{definition}

\theoremstyle{remark}

\usepackage{textcomp}
\newcommand{\pprwpm}{\text{PP}\textsubscript{RW\textpm}}

\newcommand{\ppaaa}
{\text{PP}_{3\alpha}}

\icmltitlerunning{In-context learning agents are asymmetric belief updaters}

\begin{document}

\twocolumn[
\icmltitle{In-context learning agents are asymmetric belief updaters}


\icmlsetsymbol{equal}{*}

\begin{icmlauthorlist}
\icmlauthor{Johannes A. Schubert}{mpi}
\icmlauthor{Akshay K. Jagadish}{mpi,cch}
\icmlauthor{Marcel Binz}{equal,mpi,cch}
\icmlauthor{Eric Schulz}{equal,mpi,cch}
\end{icmlauthorlist}

\icmlaffiliation{mpi}{Computational Principles of Intelligence Lab, Max Planck Institute for Biological Cybernetics, Tübingen, Germany}
\icmlaffiliation{cch}{Institute for Human-Centered AI, Helmholtz Computational Health Center, Munich, Germany}

\icmlcorrespondingauthor{Johannes A. Schubert}{johannes.schubert@tue.mpg.de}

\icmlkeywords{Machine Learning, ICML, Large Language Models, Rational Analysis, In-Context Learning, Meta Learning}

\vskip 0.3in
]



\printAffiliationsAndNotice{\icmlEqualContribution} 

\begin{abstract} 
We study the in-context learning dynamics of large language models (LLMs) using three instrumental learning tasks adapted from cognitive psychology. We find that LLMs update their beliefs in an asymmetric manner and learn more from better-than-expected outcomes than from worse-than-expected ones. Furthermore, we show that this effect reverses when learning about counterfactual feedback and disappears when no agency is implied. We corroborate these findings by investigating idealized in-context learning agents derived through meta-reinforcement learning, where we observe similar patterns. Taken together, our results contribute to our understanding of how in-context learning works by highlighting that the framing of a problem significantly influences how learning occurs, a phenomenon also observed in human cognition.
\end{abstract}

\section{Introduction}
\label{introduction}

    \begin{figure*}
        \centering
        \includegraphics[]{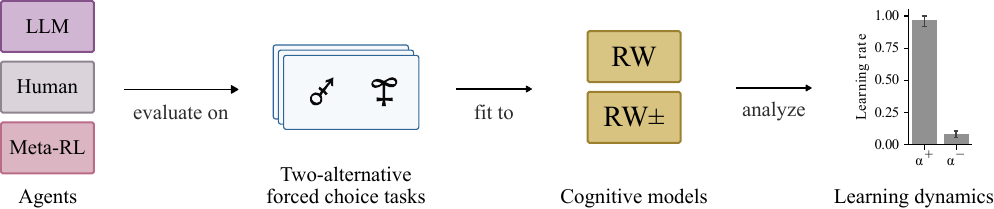}
        \caption{Schematic of our methodology, where we evaluate the learning dynamics of LLMs, humans, and meta-reinforcement learning (Meta-RL) agents on two-alternative forced choice tasks. After evaluating the agents on the tasks, we fit variants of cognitive models based on the Rescorla-Wagner (RW) model to the resulting behavior. Finally, we analyze the fitted models and extract and compare the learning rates.}
        \label{fig:overview}
    \end{figure*}


Large language models (LLMs) are powerful artificial systems that excel at many tasks \cite{radford2019language}. They can, among other things, write code \cite{roziere2023code}, help to translate from one language to another \cite{kocmi2023large}, and play computer games \cite{wang2023voyager}. Their abilities are so far-reaching that some \cite{bubeck2023sparks} have argued that they \say{could reasonably be viewed as an early (yet still incomplete) version of an artificial general intelligence (AGI) system}. At the same time, they are notoriously difficult to interpret which becomes especially aggravating as these models permeate through our society.

In the present paper, we aim to shed light on the in-context learning abilities of LLMs \cite{brown2020language}. Making use of the two-alternative forced choice (2AFC; \citealp{fechner1860elemente}) paradigm from cognitive science, we show that in-context learning implements an asymmetric updating rule when learning about the values of options. In particular, we find that -- when provided with outcomes from freely chosen options -- in-context learning exhibits an optimism bias \cite{sharot2011}, meaning that it learns more from positive than from negative prediction errors. We additionally find that this effect is mediated by two factors. First, when the outcome of the unchosen option is also provided, the bias for that option reverses and the model learns more from negative than from positive prediction errors. Furthermore, when no agency is implied and the query says \textit{someone else} does the sampling instead of \textit{you} sampled, the bias disappears. Interestingly, similar behavioral effects have been observed in human subjects \cite{lefebvre2017, chambon2020}.

Why do these tendencies for asymmetric belief updating emerge in both natural and artificial agents? Previous work has suggested that asymmetric belief updating might be a rational strategy to implement as it allows agents to achieve maximum rewards in a given task. However, these claims have been limited by the use of a restricted model class \cite{lefebvre2022, caze2013adaptive}. To investigate this idea further, we study the behavior of idealized in-context learning agents trained specifically to solve 2AFC tasks using meta-reinforcement learning (Meta-RL). Meta-RL agents have been shown to implement Bayes-optimal learning strategies upon convergence \cite{ortega2019meta, binz2023} and enable us to test if displaying such a bias is rational. We again find the same behavioral effects in these agents: (1) they show an optimism bias when only observing outcomes of the chosen option, (2) the bias reverses when learning about the value of the unchosen option, and (3) it disappears when the agent has no control about its own choices. 

Taken together, our results have broad implications for both natural and artificial agents. We have shown that in-context learning depends critically on how the problem is framed. There are many applications where practitioners have control over problem framing, and thus our results suggest that these design choices must be carefully considered to achieve desired outcomes. In the context of human cognition, our simulations extend previous work suggesting that the optimism bias (and related effects) may not be a bias after all, as it can be considered a rational response to certain problems.

\section{Related work}
    
    \textbf{In-context learning in LLMs.} 
    In their seminal paper introducing GPT-3, \citet{brown2020language} demonstrated that LLMs can do tasks, such as text translation, question-answering, and arithmetic problems, after seeing just a few demonstrations -- an ability they referred to as in-context learning.
    There has been a recent surge of research trying to better understand this phenomenon: investigating when and how in-context learning emerges \cite{chan2022data, min2022rethinking, wei2023larger}, identifying what algorithms LLMs implement during in-context learning \cite{xie2021explanation, von2023transformers}, and mapping out what can be learned in-context \cite{garg2022can, dong2022survey}. For the present paper, the work of \citet{binz2023using} and \citet{coda2023meta} is of particular relevance. They showed that LLMs can learn to perform simple multi-armed bandit problems -- which were similar to the 2AFC problems used here -- in-context and without requiring any weight updates.
    
    \textbf{Asymmertric belief updating in cognitive science.}  How humans integrate task-relevant information to update their beliefs has received a lot of attention in cognitive science \cite{jacobs2011bayesian, nassar2010approximately, gershman2015unifying}. Traditionally, this question has been investigated using the 2AFC paradigm, which provides a controlled setup to study the various facets of human reinforcement learning, including generalization, exploration, and compositional inference \cite{jagadish2023, binz2022modeling, lefebvre2017, behrens2007, chambon2020, palminteri2022, schulz2019}. 
    
    The experimental paradigms and analyses used in this paper are heavily inspired by the following studies. \citet{lefebvre2017} showed that people have asymmetric belief updating tendencies in a reinforcement learning setting. Later on, \citet{chambon2020} demonstrated that this tendency reverses when people observe outcomes for unchosen options and that it completely disappears in purely observational trials (i.e. when participants observe outcomes following a predetermined choice). Recently, \citet{palminteri2022} reviewed the influence of factors, such as reward magnitude \cite{lefebvre2022} and volatility \cite{gagne2020, behrens2007}, on asymmetric belief updating in people. Finally, \citet{lefebvre2022} used simulations to show that asymmetric belief updating can lead to optimal performance under certain reward regimes.

\section{Methods}

In this section, we first describe how we queried an LLM to perform 2AFC tasks and explain how models from cognitive psychology can be used to analyze the in-context learning dynamics of LLMs and idealized agents derived through Meta-RL. We provide an overview of our methodology in Figure \ref{fig:overview}.


\subsection{LLM prompting}

In a 2AFC task, an agent has to repeatedly choose between two options and receives a reward after each choice. The goal of the agent is to maximize the reward over all trials. We prompted an LLM to perform such 2AFC tasks. We used \textsc{Claude v1.3} as the reference LLM for all our experiments via its API\footnote{\url{https://www.anthropic.com/news/introducing-claude}} with the temperature set to 0.0. 

The prompt design was based on earlier work that had studied LLMs in similar settings \cite{binz2023using, coda2023meta}. Each prompt included an introduction to the task setup, a history of previous observations, and a question asking for the next choice. The tasks were framed in a gambling context, where an agent visits several casinos with two slot machines. 

The following prompt was used to run the default 2AFC task (see Section~\ref{sec:4}): 

\begin{tcolorbox}[ rounded corners, colback=mBlue!5!white,colframe=mBlue!75!black, title=\textbf{Prompt Task 1}]
You are going to visit four diﬀerent casinos (named 1, 2, 3, and 4) 24 times each. Each casino owns two slot machines which all return either 0.5 or 0 dollars stochastically with diﬀerent reward probabilities. Your goal is to maximize the sum of received dollars within 96 visits.\\[0.2cm]
You have received the following amount of dollars when playing in the past:\\[0.2cm] 
- Machine B in Casino 4 delivered 0.5 dollars.\\
- Machine F in Casino 1 delivered 0.0 dollars.\\
- Machine B in Casino 4 delivered 0.5 dollars.\\[0.2cm]
Q: You are now in visit 4 playing in Casino 4. Which machine do you choose between Machine R and Machine B?\\[0.2cm]
A: Machine [insert]
\end{tcolorbox}

Prompts were updated dynamically after every trial. Slot machines were labeled with a random letter, excluding meaningful ones (U, I), and the order of slot machine labels was randomized. The selected slot machine returned a stochastic reward that was appended to the bulleted history of previous slot machine interactions in subsequent prompts.    
We simulated the behavior of the LLM for a certain number of trials and recorded the action-reward pairs. 

We additionally considered two other variants of 2AFC tasks, which shared the same general structure but differed in how information was conveyed to the agent. One provided additional information by revealing the reward for the unchosen option (see Section~\ref{sec:5}) and the other included trials in which the choice of a particular option was forced (see Section~\ref{sec:6}). 
 
\subsection{Using cognitive models to analyze in-context learning dynamics} \label{sec:model-fitting}

\begin{figure*}[t]
    \centering
    \includegraphics[]{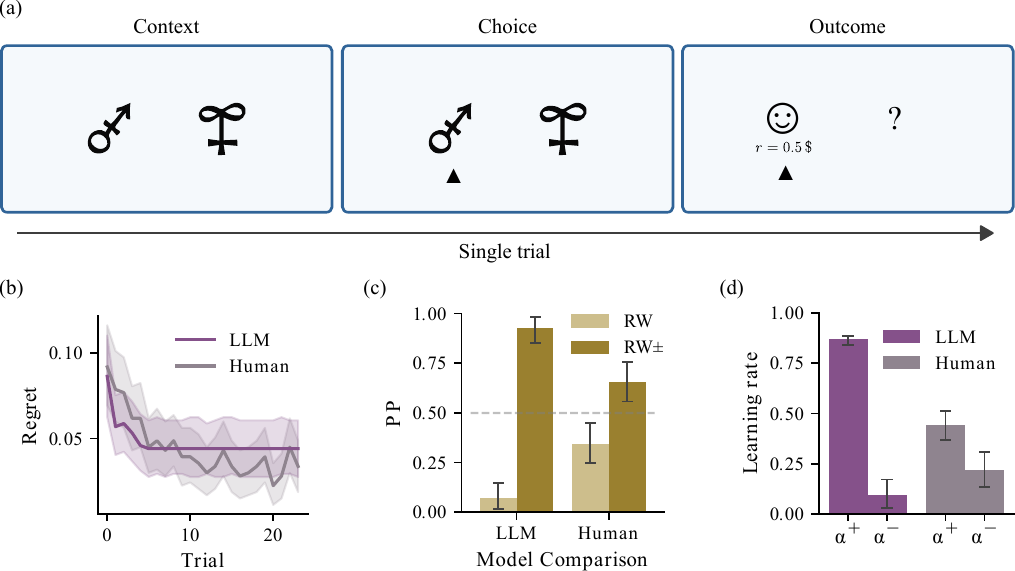}
    \caption{2AFC task with partial feedback. (a) Presentation of a single trial. First, two slot machines, shown as symbols, are presented. After a choice is made, the outcome is shown. (b) Average performance of the LLM and humans measured in terms of regret. Performance improves over trials. (c) Model comparison of the Rescorla-Wagner (RW) model and the RW$\pm$ model. For both the LLM (left) and human participants (right), RW$\pm$ provides a better fit to the data, as indicated by the average posterior probability (PP). (d) Average learning rates of the RW$\pm$ model for the LLM and human participants. Both agents show a stronger response to positive prediction errors than to negative prediction errors. Human participant data reproduced from \cite{lefebvre2017}. Error bars and shaded areas and correspond to 95\% CIs.}
    \label{fig:partial-task}
\end{figure*}

For the analysis of the learning dynamics of an agent, we built simplified but interpretable cognitive models of its choice behavior. We then identified which parameter setting in these models provides the best explanation for the observed data. The resulting parameter values can then offer a window into the behavior of the agent. This approach has its origins in cognitive science \cite{miller1956magical, rescorla1972, wilson2019} but has recently been adopted to study the behavior of artificial agents as well \cite{dasgupta2022language, binz2023using, bigelow2023context}.

The core of our analysis is the Rescorla-Wagner (RW) model \cite{rescorla1972} -- a classic model for studying learning in humans. It formalizes learning as a dynamic process of minimizing prediction errors between expected and actual outcomes:
\begin{align*}
    V_{t+1}(a) &= V_t(a) + \alpha \cdot \delta_t \\
    \delta_t &= r_t - V_t(a)
\end{align*}
where $\delta_t$ is the prediction error between the observed reward $r_t$ and the expected reward value $V_t(a)$ in trial $t$. The learning rate $\alpha$ determines how much the expected value for action $a$ changes after an observation. Thus, it quantifies the amount of learning that occurs based on the prediction error. The model maps learned values to choice probabilities using a softmax decision rule with an inverse temperature parameter $\beta$:
\begin{equation*}
    p(a)=\frac{\exp(\beta \cdot V_{t}(a))}{\sum_{k=1}^{K}\exp(\beta \cdot V_{t}(k))}
\end{equation*}

There are several extensions to the RW model that have been used to evaluate different hypotheses about how learning occurs. We relied on one such extension, namely the RW$\pm$ model \cite{palminteri2022}, which introduces separate learning rates for positive and negative prediction errors: 
\begin{align*}
    V_{t+1}(a)&=V_t(a)+\left\{\begin{array}{ll}
    \alpha^+ \cdot \delta_t, & \text { if } \delta_t>0 \\
    \alpha^- \cdot \delta_t, & \text { if } \delta_t<0
    \end{array}\right. \\
    \delta_t &= r_t - V_t(a) \nonumber
\end{align*}
Positive prediction errors occur in situations where the received reward is greater than the estimated value (i.e. $\delta_t > 0$), while negative prediction errors occur when the received reward is less than the estimated value (i.e. $\delta_t < 0$). 

This model allows us to study how the information of prediction errors is weighted during learning. In the case where the expected value is influenced symmetrically by both prediction errors (i.e. $\alpha^{+}$ = $\alpha^{-}$), the RW$\pm$ model is equivalent to the classical RW model. If the learning rates differ, the beliefs are updated asymmetrically, either optimistically ($\alpha^+ > \alpha^-$) or pessimistically ($\alpha^- > \alpha^+$). 

We relied on a maximum a posteriori estimation approach to fit the parameters of these models to the behavioral data generated by an LLM. Prior probabilities were based on \citet{daw2011model}, using a beta distribution $\mathcal{B}(1.1, 1.1)$ for learning rates and a gamma distribution with $\mathcal{G}(1.2, 5)$ for the inverse temperature parameter. Each option was represented by a separate expected value $V_t$ and the initial values $V_0$ were assigned to the average reward values.

For each task, we compared different cognitive models based on their average posterior probabilities (PP; \citealp{wu2020simple}). To do this, we first computed the Bayesian Information Criterion (BIC; \citealp{schwarz1978}) for each model $m$ and each simulation:
\begin{equation*}
    \text{BIC}_m = k \cdot \ln(N) - 2 \cdot \ln p(a_{1:N} | \hat{\theta}, m)
\end{equation*}
where $k$ is the number of parameters, $N$ is the number of choices performed, and $\hat{\theta}$ are the estimated parameters based on the agent's choices. Under the assumption of a uniform prior over models, the PP for each simulation can then be approximated as:
\begin{align*}
    p(m | a_{1:N}) \approx& \frac{\exp(-0.5 \cdot \text{BIC}_m)}{\sum_i \exp(-0.5 \cdot \text{BIC}_i)}
\end{align*}
    
\section{Partial information results in optimism bias} \label{sec:4}

\begin{figure*}[!ht]
    \centering
    \includegraphics[]{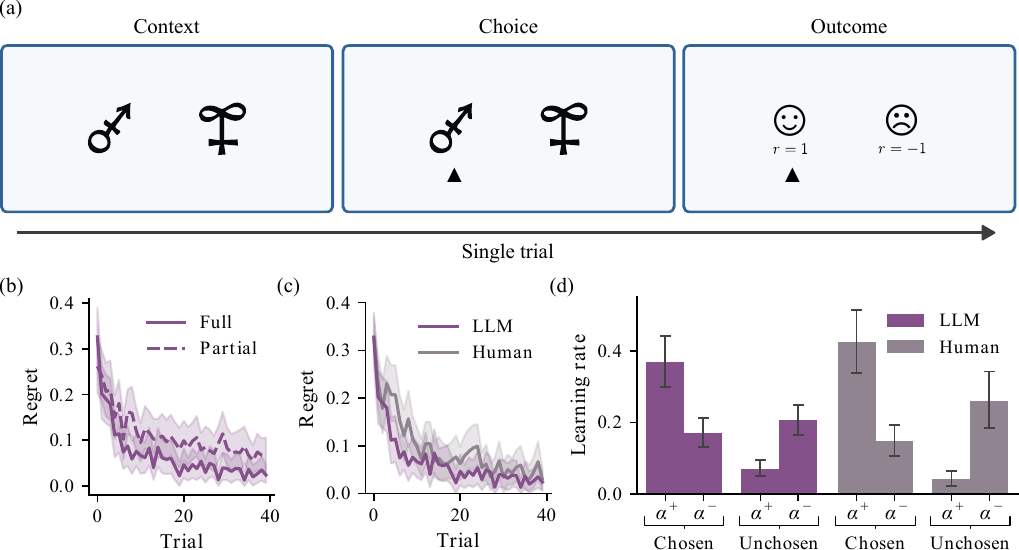}
    \caption{2AFC task with full feedback. (a) Presentation of a single trial: The two slow machines are again shown as symbols. After a choice is made, the outcome of both the chosen and the unchosen option is shown. (b) Average regret of the LLM on partial and full feedback blocks, showing that the additional information of full feedback blocks leads to improved performance. (c) Average regret of the LLM and humans for full feedback blocks. The performance of the LLM improves over trials and is on par with human performance. (d) Learning rates of the full feedback model with two learning rates -- for positive and negative prediction errors -- for the chosen and unchosen slot machine. Both agents have an optimism bias for the chosen option and a pessimism bias for the unchosen option. Human participant data reproduced from \cite{chambon2020}. Error bars and shaded areas correspond to 95\% CIs.}
    \label{fig:full-task}
\end{figure*}

We started our investigations with the 2AFC task with default settings in which only the outcome for the chosen slot machine was revealed as illustrated in the prompt of the previous section and in Figure~\ref{fig:partial-task}a. We adopted the experimental paradigm of an earlier study with humans from \citet{lefebvre2017}. It involved four different casinos with win probabilities of 0.25/0.25, 0.25/0.75, 0.75/0.25, and 0.75/0.75 for the respective slot machines. Winning led to a reward of $0.5$ dollars, losing to a reward of $0.0$ dollars. Each casino was visited $24$ times in a random order resulting in $96$ visits in total. We simulated the LLM 50 times on the task. 

First, we examined performance regarding regret, which is the amount of reward missed relative to the optimal choice. When averaged across all simulations and casinos, regret decreased significantly from $0.09 \pm 0.01$ in the first trial to $0.05 \pm 0.01$ in the last trial ($t(199) = 3.1, p = .002$; see Figure~\ref{fig:partial-task}b). In terms of performance, the improvement during in-context learning is similar to that observed in human studies \cite{lefebvre2017}. 

To investigate whether in-context learning is done symmetrically or asymmetrically, we fitted the classical RW and the RW$\pm$ model to the simulated behavior of the LLM separately for each simulated run as described in Section~\ref{sec:model-fitting}. The model comparison indicated that the RW$\pm$ model provided on average a better explanation of the data with a PP of $0.93 \pm 0.04$ (see Figure~\ref{fig:partial-task}c). Analyzing the learning rates revealed that this was associated with a strong optimistic asymmetry: new information about the options was incorporated more readily when it was desirable (positive prediction errors) than undesirable (negative prediction errors) as shown in Figure~\ref{fig:partial-task}d, with $\alpha^+ > \alpha^-$ ($\alpha^+ = 0.87 \pm 0.01, \alpha^- = 0.10 \pm 0.04$; $t(49)=20.1, p < .0001$). In-context learning seems to overweigh new evidence that conveys a positive valence. Previous work with human subjects has observed a similar -- although less pronounced -- optimism bias (reproduced in Figure~\ref{fig:partial-task}d for reference).

\section{Pessimistic updating for unchosen options} \label{sec:5}

\begin{figure*}[t]
    \centering
    \includegraphics[]{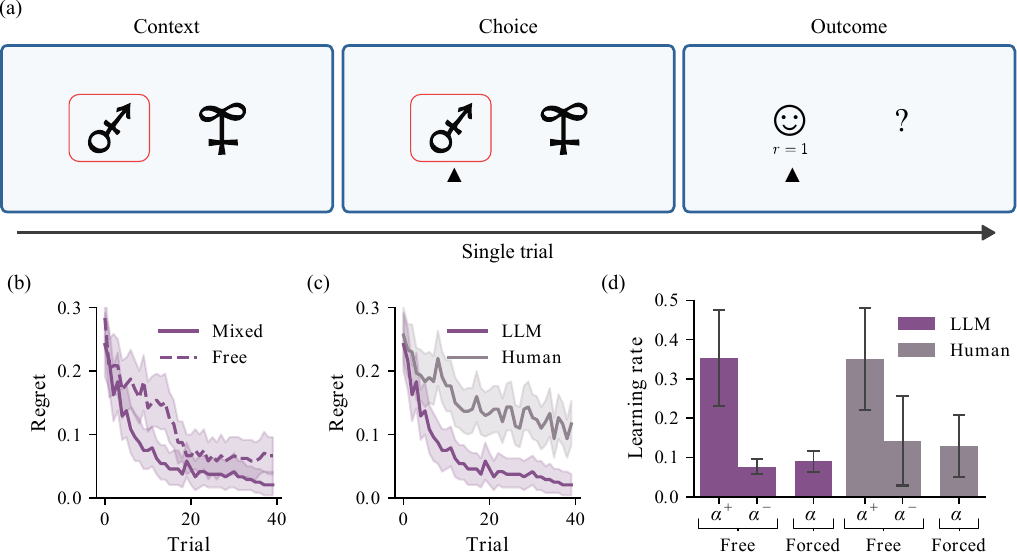}
    \caption{2AFC task for the agency condition (a) Presentation of a single forced-choice trial: In forced-choice trials, one of the two slot machines is preselected (red square) and its outcome is presented directly to the LLM. (b) Average regret of the LLM in mixed-choice and free-choice blocks, showing that the additional information of forced-choice trials in mixed-choice blocks leads to improved performance. (c) Average regret for the LLM and humans for mixed-choice blocks. The performance of the LLM outperforms the human participants over trials. (d) Learning rates of the $3\alpha$ model with two learning rates for the free-choice trials and one learning rate for the forced-choice trials. Both agents integrate feedback for positive and negative prediction errors in free-choice trials asymmetrically, whereas feedback from forced-choice trials is integrated symmetrically. Human participant data reproduced from \cite{chambon2020}. Error bars and shaded areas correspond to 95\% CIs.}
    \label{fig:agency-task}
\end{figure*}

Next, we investigated the in-context learning dynamics of LLMs in a setting that provided full feedback about the outcomes of both the chosen and unchosen slot machines. For this, we borrowed another experimental paradigm of an earlier study with humans from  \citet{chambon2020}. The adapted task consisted of multiple blocks, each containing only a single casino and two slot machines. We prompted each of these blocks independently. Half of the blocks provided full feedback, the other half only provided partial feedback about the chosen machine.\footnote{Note that the task also contained forced-choice trials in which the agent has to select a predetermined machine. We ignored these trials for the analysis presented in this section, but come back to them in the next section.} In the full feedback blocks, the foregone outcome of the unchosen slot machine was provided in addition to the outcome of the chosen slot machine in the history of prior interactions (see Figure~\ref{fig:full-task}a). We adapted the feedback items as follows to reflect this change: \say{On visit 1 you played Machine H and earned 1.0 point. On Machine E you would have earned -1.0 point.}

Half of the blocks were high-reward blocks with reward probabilities of 0.9 and 0.6 for both machines, and the other half were low-reward blocks with reward probabilities of 0.4 and 0.1. In all blocks, the outcome resulted in either a gain or a loss of one point. The prompt in the full feedback blocks also mentioned that the observed outcome of the unchosen machine would not be added to the total rewards earned (see Appendix~\ref{app:full-prompt}). We simulated the LLM 24 times on the task consisting of 16 blocks with 40 trials each.

When comparing the partial to the full feedback blocks, we saw a decrease in final regret from partial feedback ($0.07 \pm 0.01$) to full feedback blocks ($0.02 \pm 0.01$), with $t(191) = 2.9, p=.004$ (see Figure~\ref{fig:full-task}b). This suggests that in-context learning was able to incorporate the additional information conveyed by the unchosen option. Like in the previous section, performance in terms of regret was comparable to that observed in an earlier study with human subjects as shown in Figure~\ref{fig:full-task}c.

For analyzing the dynamics of in-context learning, we extended the RW$\pm$ model to also account for the additional information provided by the unchosen option. This extended RW$\pm$ model included separate learning rates for positive and negative prediction errors of the chosen and unchosen options (i.e. four learning rates in total).

Figure~\ref{fig:full-task}d illustrates that fitted learning rates for chosen and unchosen options show opposite asymmetric patterns. While we again observed an optimism bias for learning about the chosen machine ($t(23)=5.9, p<.0001$), information for the unchosen machine was integrated such that negative prediction errors were preferentially taken into account, relative to positive ones ($t(23)=5.7, p<.0001$). This pattern is known as confirmation bias as it refers to integrating information in a way that confirms prior beliefs \cite{palminteri2017, nickerson1998confirmation}. Earlier studies with people in a matched experiment \cite{chambon2020} found similar behavioral characteristics (see Figure~\ref{fig:full-task}d for reference). 

The pattern of learning rates suggests that the four learning rates can be compressed into just two: a confirmatory learning rate that combines the positive chosen and negative unchosen learning rates and a disconfirmatory learning rate that combines the negative chosen and positive unchosen learning rates. We therefore fitted a second cognitive model that was obtained by merging the learning rates depending on the confirmatory or disconfirmatory outcome. This simplified model provided a better fit to the data with an average PP of $0.92 \pm 0.03$ (refer to Appendix~\ref{fig:full-confirmation-model} for the learning rates of this model). This implies that LLMs update their beliefs about certain outcomes more when new evidence confirms their prior beliefs and past decisions than when it disconfirms or contradicts them.

\section{No asymmetric updating without agency} \label{sec:6}

In our final analysis, we examined the influence of agency on the in-context learning dynamics of LLMs by providing additional information about slot machines through observational trials. We used the same general task structure as in the previous section \cite{chambon2020}. However, half of the blocks now included randomly interleaved forced-choice trials of another agent playing in the casino (mixed-choice blocks; see Figure~\ref{fig:agency-task}a). The other half contained only free-choice trials to assess the performance improvements resulting from the additional information provided by the forced-choice trials in the mixed-choice blocks. Both types of blocks provided partial feedback about the outcome of the selected machine. 

We adapted the prompt structure of the force-choice trials as follows: \say{On visit 1 someone else played Machine H and received 1.0 point}. To avoid biasing the agent towards a particular machine, the forced-choice trials sampled both slot machines equally. The prompt of the mixed-choice blocks mentioned that rewards from forced-choice trials would not be added to the total reward (see Appendix~\ref{app:agency-prompt}). As in the previous section, half of the blocks were high-reward blocks, the other half were low-reward blocks. The six mixed-choice blocks consisted of 80 trials with 40 free and 40 forced-choice trials, while the six free-choice blocks consisted of only 40 free-choice trials. We simulated the LLM on this task 24 times. 

When comparing the free-choice blocks with the mixed-choice blocks, we found that the LLM incorporated the additional information from the forced-choice trials, leading to performance improvements with a decrease in regret from $0.07 \pm 0.2$ to $0.02 \pm 0.01$ ($t(143)=2.5, p=0.01$; see Figure~\ref{fig:agency-task}b). In comparison to human data from an earlier study \cite{chambon2020}, the LLM learned significantly faster in this setting as shown in Figure~\ref{fig:agency-task}c (final regret for LLMs: $0.02 \pm 0.01$, final regret for humans: $0.07 \pm 0.02$; $t(143)= 2.7, p=.008$). 

To analyze the effects of choice types on learning we fitted two different cognitive models to the behavior of the LLM in the mixed-choice blocks: (1) a $4\alpha$ model which consisted of separate learning rates for positive and negative prediction errors for both free-choice and forced-choice trials and (2) a $3\alpha$ model which consisted of separate learning rates for positive and negative prediction errors for free-choice trials and only one learning rate for forced-choice trials. The model comparison indicated that the $3\alpha$ model represented the behavior of the LLM best ($\ppaaa = 0.77 \pm 0.06$; see Figure~\ref{fig:agency-task}c), suggesting that it seems to integrate the information from forced-choice trials symmetrically (i.e. $\alpha^{+}$ = $\alpha^{-}$). In contrast, information from free-choice trials is asymmetrically weighted with $\alpha^{+} = 0.35 \pm 0.06$ being greater than $\alpha^{-} = 0.08 \pm 0.01$ ($t(23)=4.7, p = .0001$), as seen in Figure~\ref{fig:agency-task}d.
This implies the extent to which the LLM can control its environment changes how it integrates received information. 

\section{Idealized in-context learning agents also display asymmetric updating} \label{sec:meta-rl}

\begin{figure*}[t]
    \centering
    \includegraphics[]{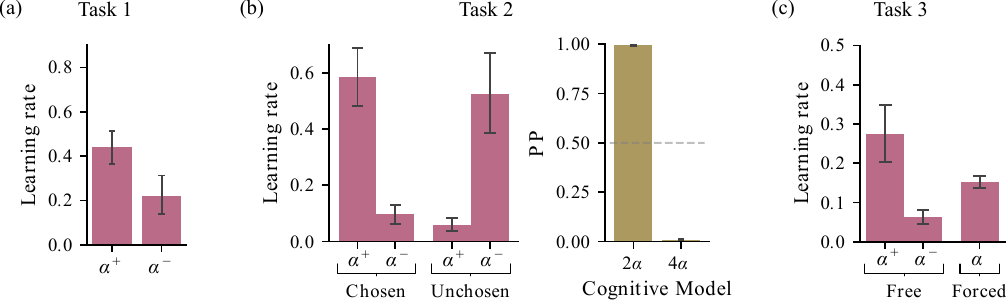}
    \caption{Learning rate analyses for the Meta-RL agent. (a) In the partial feedback task, the RW$\pm$ provided a better fit to the Meta-RL agents' behavior ($\pprwpm= 0.99 \pm 0.00$) and showed an optimistic tendency to integrate information. (b) In the full feedback task, the tendency to integrate positive outcomes for chosen options optimistically and negative outcomes for unchosen options pessimistically is even more pronounced than in the LLMs (left). Model comparison showed that the simplified confirmatory model ($2\alpha$) fits the data better (right). (c) In the agency condition, the $3\alpha$ best fit the simulated behavior ($\ppaaa=0.85 \pm 0.04$), implying that information was integrated asymmetrically in free-choice trials and symmetrically in forced-choice trials. Error bars correspond to 95\% CIs.}
    \label{fig:meta-rl-learning-rates}
\end{figure*}

To better understand why in-context learning exhibits these behavioral characteristics, we tested whether they also emerge in a more controlled setting. For this, we trained idealized transformer-based agents to solve our previously examined 2AFC tasks via in-context learning. The agent received the previously selected action $a_{t-1}$ and the reward that followed $r_{t-1}$, alongside the current context $c_t$ which varied slightly for each task (see Appendix~\ref{app:meta-rl-agent}), as inputs in each time-step. We trained the agent using Meta-RL \cite{duan2016rl, wang2016learning} to learn a history-dependent policy $\pi_{\mathbf{\Theta}}$ that maximizes the expected sum of rewards over a prespecified task distribution:
\begin{equation*} \label{eq:rlobj} 
  \max_{\mathbf{\Theta}}~ \mathbb{E}_{p(\mathbf{\theta}, c_{1:T})\prod p(r_t | a_t, c_t, \mathbf{\theta}) \pi_{\mathbf{\Theta}}(a_t | a_{1:t-1}, r_{1:t-1}, c_{1:t})} \left[ \sum_{t=1}^T r_t \right]
\end{equation*}

where $a_{1:t-1}$, $r_{1:t-1}$, and $c_{1:t}$ denote sequences of actions, rewards, and contexts respectively, while $\mathbf{\Theta}$ denotes the weights of the underlying transformer. The task distribution used for training is specified by $p(\mathbf{\theta}, c_{1:T})$ and $p(r_t | a_t, c_t, \theta)$ with $\mathbf{\theta}$ corresponding to a vector containing win probabilities for all slot machines.

The distribution shared a similar structure in all three experiments. We assumed that win probabilities for each option were sampled from a uniform distribution at the beginning of each training episode to capture the assumption of an uninformative prior. The agent consisted of a Transformer network \cite{vaswani2017attention} with a model dimension of $8\cdot\mathrm{input\_size}$, two feedforward layers with a dimension of 128, and eight attention heads, followed by two linear layers that output a policy and a value estimate, respectively. The network weights were adjusted by gradient-based optimization using \textsc{ADAM} \cite{kingma2014} on a standard actor-critic loss \cite{mnih2016} at the end of each training episode. Further details about the training are provided in Appendix~\ref{app:meta-rl-sec}.

We trained all Meta-RL agents until convergence and then tested their in-context learning abilities without performing any further updates to the network weights. It has been shown that the history-dependent policy learned in this setting can solve new but similar tasks in an approximately Bayes-optimal way \cite{mikulik2020, ortega2019meta}, therefore allowing us to investigate belief updating in an idealized setting.

We simulated the agents on all three tasks and analyzed their behavior as described in the previous sections. We found that the idealized transformer-based agents learned strategies that outperformed both LLMs and humans as shown in Appendix~\ref{fig:meta-rl-perf}. Furthermore, the idealized agents showed similar learning characteristics to LLMs: (1) in the partial feedback task, they learned more from positive prediction errors, (2) the pattern reversed for the counterfactual option in the case of the full feedback task, and (3) asymmetric updating was limited to free-choice trials and was absent in forced-choice trials. These results are illustrated in Figure~\ref{fig:meta-rl-learning-rates}. 

Taken together, these results indicate that behavior observed in humans and LLM shares key characteristics with idealized in-context learning agents trained specifically on 2AFC tasks. 

\section{Discussion}

People change their learning strategies based on how the problem is framed \cite{palminteri2022}. In this paper, we have shown that this also holds for in-context learning agents. In particular, we found that LLMs exhibit an optimism bias, i.e. they learn more from better-than-expected outcomes (positive prediction errors) than from worse-than-expected ones (negative prediction errors). However, this bias was only present when the prompt was formulated in a way that implied agency. Furthermore, we found that for counterfactual feedback for unchosen options, the bias reversed and the model learned more from negative than positive errors for these options. 

We conducted these analyses in a highly controlled setting, providing high internal validity to our results. However, claims regarding their external validity must be taken with care for now, and future studies will have to investigate whether we can also find similar patterns in more naturalistic settings. Furthermore, our findings relied on an inference from observed learning rate patterns in cognitive models. It is unclear if the link is causal as another auxiliary computational process could potentially be explaining the observed pattern. Nevertheless, opposing interpretations of the pattern have not yet been substantiated by experimental evidence in human experiments \cite{palminteri2022}. 

In the tasks we have investigated, it is rational to perform asymmetric belief updating -- as indicated by our simulations with Meta-RL agents. It remains an open question if this also holds in situations where asymmetric belief updating is suboptimal. Future work should aim to characterize whether in-context learning also displays asymmetric belief updating in such situations. The study of \citet{globig2021under} could be a starting point for this question as it provides an example of a setting where people show an optimism bias even though it is not rational.

From a methodological perspective, our work demonstrates that it is possible to fit simpler computational models to the behavior of LLMs and use the resulting parameter values to infer \emph{how} they behave. 
Fitting and interpreting parameters of simpler computational methods provides us with a tool that complements existing techniques for explainable machine learning \cite{roscher2020explainable}. We believe that there are further exciting applications in this research field.

Taken together, our results contribute to our understanding of how in-context learning in LLMs works, which is especially important as the number of applications of these models in real-world scenarios is increasing \cite{binz2023a, eloundou2023gpts, kasneci2023chatgpt}. If the biases found in this work also emerge in tasks where they are not optimal, as has been shown in humans \cite{shepperd2013}, it will be important to develop techniques to mitigate them.

\bibliography{paper}
\bibliographystyle{paper}

\newpage
\appendix
\onecolumn

\section{Task 2 additional information}

\subsection{Prompts} \label{app:full-prompt}
The full information task consisted of 16 blocks, half of which were partial feedback blocks and half of which were full feedback blocks. Below is a sample prompt for both types of blocks. The differences between the partial and full feedback blocks are shown in bold. 

\begin{tcolorbox}[ rounded corners, colback=mBlue!5!white,colframe=mBlue!75!black, title=\textbf{Prompt Task 2: Partial Feedback Blocks}]
You will visit a casino 40 times. The casino has two slot machines that stochastically return either 1 or -1 with different reward probabilities. You can only interact with one slot machine per visit. Half of the time you visit the casino, you can play, the other someone else is playing. During visits where you can play, you'll earn points from the chosen machine. During visits where someone else is playing, you'll learn what points are earned on the chosen machine. Your goal is to maximize the total amount of points you receive in all 20 visits you can play. \\[0.2cm]
During your previous visits you have observed the following:\\[0.2cm]
- On visit 1 someone else played Machine H and earned 1.0 point.\\
- On visit 2 you played Machine H and earned 1.0 point.\\
- On visit 3 you played Machine E and earned -1.0 point.\\
Q: You are now in visit 4. Which machine do you choose between Machine E and Machine H?\\
A: Machine [insert]
\end{tcolorbox}

\begin{tcolorbox}[ rounded corners, colback=mBlue!5!white,colframe=mBlue!75!black, title=\textbf{Prompt Task 2: Full Feedback Blocks}]
You will visit a casino 40 times. The casino has two slot machines that stochastically return either 1 or -1 with different reward probabilities. You can only interact with one slot machine per visit. Half of the time you visit the casino, you can play, the other someone else is playing. During visits where you, can play, you'll earn points from the chosen machine. \textbf{You'll also learn what points would have been earned had the other machine been selected.} During visits where someone else is playing, you'll learn what points are earned on the chosen \textbf{and what points would have been earned had the other machine been selected. Nevertheless, you only accumulate points from the machine you choose to play.} Your goal is to maximize the total amount of points you receive in all 20 visits you can play. \\[0.2cm]
During your previous visits you have observed the following:\\[0.2cm]
- On visit 1 someone else played Machine H and earned 1.0 point.\\
\hspace*{0.4em}\textbf{On Machine E the player would have earned -1.0 point.}\\
- On visit 2 you played Machine H and earned 1.0 point.\\
\hspace*{0.4em}\textbf{On Machine E you would have earned -1.0 point.}\\
- On visit 3 you played Machine E and earned -1.0 point.\\
\hspace*{0.4em}\textbf{On Machine H you would have earned 1.0 point.}\\[0.2cm]
Q: You are now in visit 4. Which machine do you choose between Machine E and Machine H?\\
A: Machine [insert]
\end{tcolorbox}

\subsection{Model comparison}

\begin{figure}
    \centering
    \includegraphics{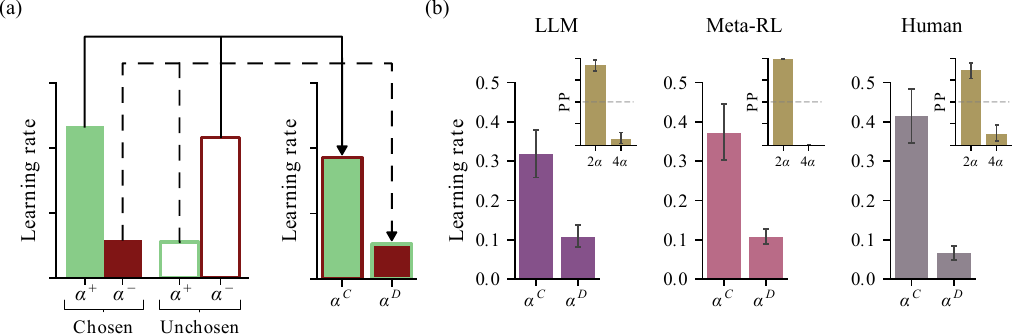}
    \caption{Confirmation bias in the full feedback task. (a) Schematic showing how the learning rates of the full model (i.e. a model for a different learning rate for each possible combination of outcome types and prediction error types) relate to those of the confirmation bias model ($2\alpha$), which bundles together the learning rates for positive chosen and negative unchosen (i.e. confirmatory) prediction errors ($\alpha^C$) and the learning rates for negative chosen and positive unchosen (i.e. disconfirmatory) prediction errors ($\alpha^D$). Adapted from \cite{palminteri2022} (b) Fitted confirmation bias model for the LLM, the Meta-RL agent and human participants. The average posterior probabilities (PP) indicate that the $2\alpha$ model is a superior fit in all model comparisons. Error bars correspond to 95\% CIs.}
    \label{fig:full-confirmation-model}
\end{figure}

We simplified the analysis of the original by fitting separate learning rates only for the chosen and unchosen option, but not separate learning rates for free and forced-choice trials. Furthermore, we used only the simulated behavior from full feedback blocks for the fitting of two cognitive models -- a $2\alpha$ and a $4\alpha$ model. 

The model comparison revealed that the $2\alpha$ model provided a better fit to the behavior. The $2\alpha$ model contained only two learning rates -- combining the learning rates for the chosen and unchosen options, which either confirmed or disconfirmed prior beliefs. As can be seen in Figure~\ref{fig:full-confirmation-model}, all agents show a clear tendency to overweight information that confirms their choices.

\section{Task 3 additional information}

\subsection{Prompts}\label{app:agency-prompt}

The agency condition consisted of 12 blocks and provided only partial feedback. Half of these blocks were free-choice blocks containing only free-choice trials, and the other half were a mixture of free-choice and forced-choice trials. 
The free-choice blocks consisted of 40 trials per block. The mixed-choice blocks consisted of 80 trials with 40 free-choice trials and 40 forced-choice trials. Below is a sample prompt for both types of blocks. The differences between the free-choice and mixed-choice blocks are shown in bold. 

\begin{tcolorbox}[ rounded corners, colback=mBlue!5!white,colframe=mBlue!75!black, title=\textbf{Prompt Task 3: Free-Choice Blocks}]
You will visit a casino 40 times. The casino has two slot machines that stochastically return either 1 or -1 with different reward probabilities. You can only interact with one slot machine per visit. Your goal is to maximize the total amount of points you receive in all 40 visits you can play.\\[0.2cm]
During your previous visits you have observed the following:\\[0.2cm]
- On visit 1 you played Machine H and earned 1.0 point.\\
- On visit 2 you played Machine N and earned -1.0 point.\\
- On visit 3 you played Machine H and earned -1.0 point. \\[0.2cm]
Q: You are now in visit 4. Which machine do you choose between Machine N and Machine H?\\
A: Machine [insert]
\end{tcolorbox}

\newpage 

\begin{tcolorbox}[ rounded corners, colback=mBlue!5!white,colframe=mBlue!75!black, title=\textbf{Prompt Task 3: Mixed-Choice Blocks}]
\textbf{You will visit a casino 80 times.} The casino has two slot machines that stochastically return either 1 or -1 with different reward probabilities. You can only interact with one slot machine per visit.\textbf{ Half of the time you visit the casino, you can play, the other half someone else is playing and you can only see the rewards for their chosen slot machine.} Your goal is to maximize the total amount of points you receive in all 40 visits you can play.\\[0.2cm]
During your previous visits you have observed the following:\\[0.2cm]
- On visit 1 you played Machine H and earned 1.0 point.\\
- \textbf{On visit 2 someone else played Machine N and received -1.0 point.}\\
- On visit 3 you played Machine H and earned -1.0 point. \\[0.2cm]
Q: You are now in visit 4. Which machine do you choose between Machine N and Machine H?\\
A: Machine [insert]
\end{tcolorbox}

\subsection{Model comparison}

\begin{figure}
    \centering
    \includegraphics{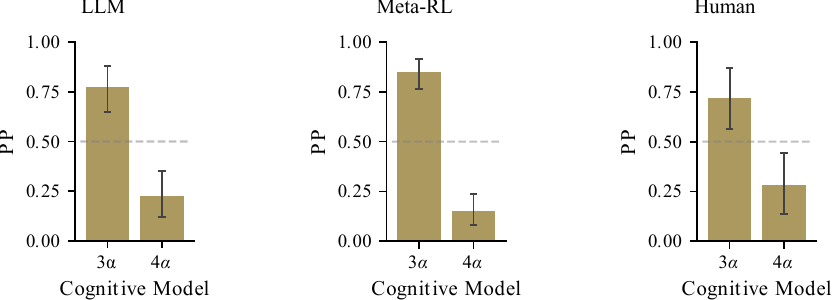}
    \caption{Model comparison for the agency condition for all agents. The $3\alpha$ model has the highest average posterior probabilities (PP). This indicates that information from forced-choice trials where no agency is involved are weighted symmetrically whereas the free-choices are weighted asymmetrically. Error bars correspond to 95\% CIs.}
    \label{fig:agency-model-comparison}
\end{figure}

We only fit two cognitive models to the mixed-choice blocks. One model consisted of separate learning rates for positive and negative prediction errors for free- and forced-choice trials ($4\alpha$). The second model consisted of two learning rates for free-choice trials and only one learning rate for forced-choice trials ($3\alpha$). Model parameters were fit based on the free-choice trials of the mixed-choice blocks. Model comparison indicated that for all agents, the $3\alpha$ model provided a better fit to the data based on the average PP (see Figure~\ref{fig:agency-model-comparison}).

\section{Meta-RL addition} \label{app:meta-rl-sec}

\subsection{Agent} \label{app:meta-rl-agent}
The agent consisted of a Transformer network with a model dimension of $8\cdot\mathrm{input\_size}$, two feedforward layers with a dimension of 128, and eight attention heads, followed by two linear layers that output a policy and a value estimate, respectively. 
The agent received the previously selected action (one-hot encoded) and the reward of that action, alongside the current context. This context included a normalized time index for all tasks. For task 1, a one-hot encoded representation of the four casinos was added, resulting in an input size of eight. The context of tasks 2 and 3 included a bit representation of the trial type (i.e. $00 =$ free-choice, $10 =$ forced-choice left option chosen, $01 = $ forced-choice right option chosen) for the current and previous trial. Task 2 additionally included the reward of the current unchosen option, resulting in an input size of eight for Task 2 and nine for Task 3. 
In the partial feedback blocks of Task 2, a placeholder of 0 for the missing reward signal of the unchosen option was propagated. 
To prevent learning from forced-choice trials, we masked the policy and value loss for those trials. To ensure a consistent input dimension, we use placeholder values for the initial inputs and all the unseen trials.

Network weights were adjusted by gradient-based optimization using \textsc{ADAM} \cite{kingma2014} on a standard actor-critic loss \cite{mnih2016} at the end of each training episode. The initial learning rate for \textsc{ADAM} was $0.0003$. 
For the actor-critic loss, we used a discount factor of $0.8$ and weighted the critic loss with 0.5. In addition, we used entropy regularization to discourage premature convergence to sub-optimal policies \cite{williams1991a}. Starting with an entropy coefficient of 1, we linearly decayed the influence of the entropy term to 0 after half of the $5,000$ episodes. We used a batch size of 64 during training. 

\subsection{Training} \label{app:meta-rl-training}
\begin{figure}[t!]
    \centering
    \includegraphics[width=1\textwidth]{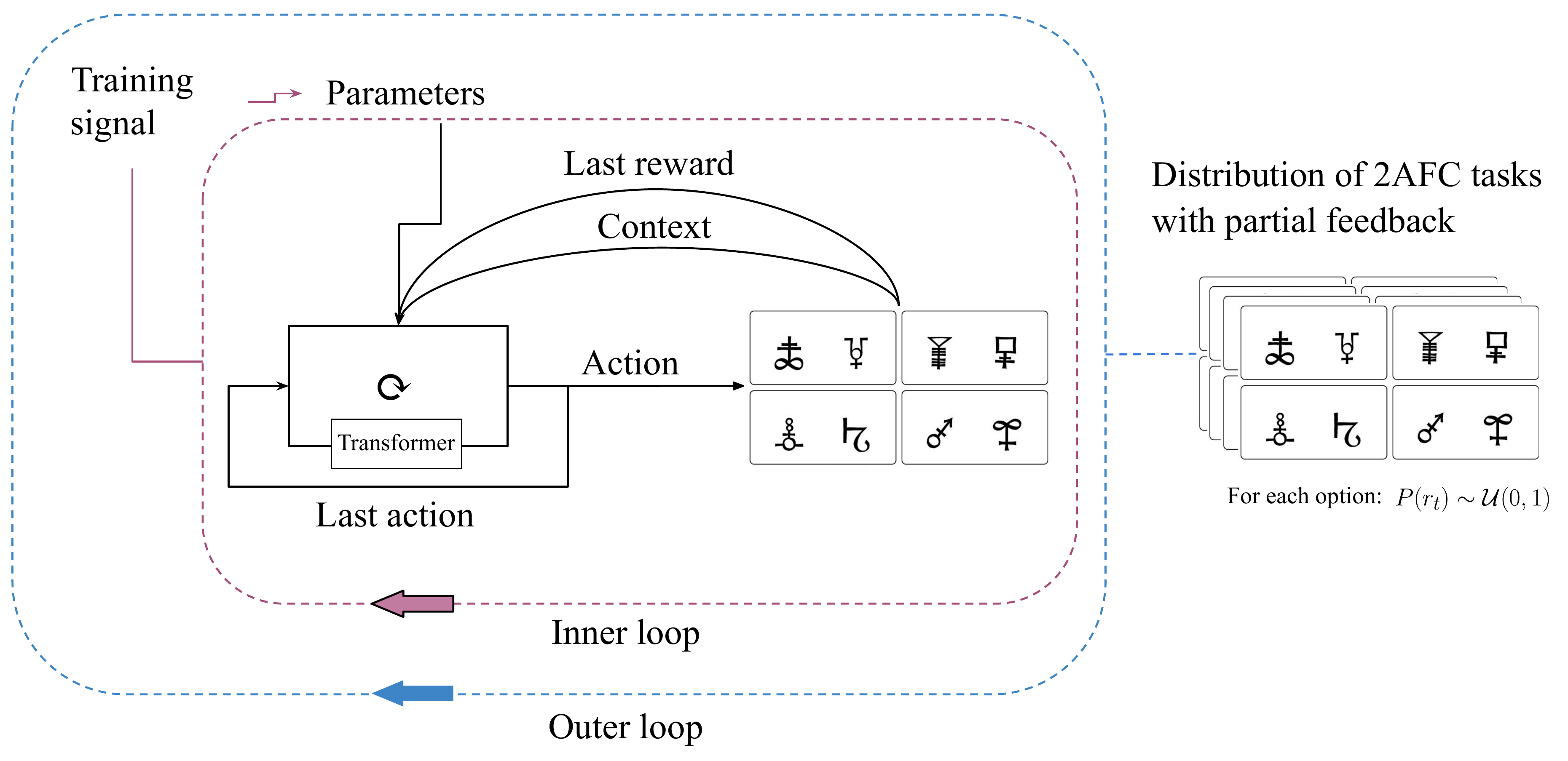}
    \caption{Schematic of the Meta-RL training for Task 1 highlighting the inner and outer training loops. The outer loop adjusts the weights of the Transformer in response to the learning experience. These weights shape the behavior of the Transformer in the inner loop as it interacts with 2AFC tasks (here Task 1). After each cycle of the outer loop, a new task is sampled where the probabilities for each option are sampled from a uniform distribution. Adapted from \citet{botvinick2019}.}
    \label{fig:training-meta-rl}
\end{figure}

The training process of the Meta-RL agent is graphically depicted in Figure~\ref{fig:training-meta-rl}. In the process, there are two optimization loops -- an outer and an inner loop. 
At the beginning of each training episode (outer loop), we sample a new task $b_{i}$ from the prior distribution of 2AFC tasks. As stated earlier, we assumed a uniform distribution for the win probabilities for each option in each task. The goal of the agent is to find a history-dependent policy $\pi_{\mathbf{\Theta}}$ that maximizes the expected total discounted reward accumulated during an episode. For that, the parameters $\mathbf{\Theta}$ are adjusted at the end of each episode. Since the decision strategy changes across training episodes, the agent must act differently according to its prior belief over which part of the task distribution it is currently in. As the optimization maximizes the expected total rewards across tasks, the policy starts to generalize the underlying principles that help reach this objective.

The agent interacts in the inner loop with the specific sampled task $b_i$, aiming to maximize its rewards across all steps with the help of its policy. At the beginning of each step, a context $c_{t}$ is drawn from a uniform distribution. Upon receiving an action $a_{t}$, the environment computes a reward $r_{t}$ and samples the next context $c_{t+1}$ to which the agent steps forward. The next context $c_{t+1}$, the action $a_{t}$, and the reward $r_{t}$ are concatenated and added to the input to the Transformer. As demonstrated by \citet{wang2017}, this input design is crucial for an agent to learn an association between choices that have been made in particular states and their subsequent rewards.

\begin{figure}
    \centering
    \includegraphics[]{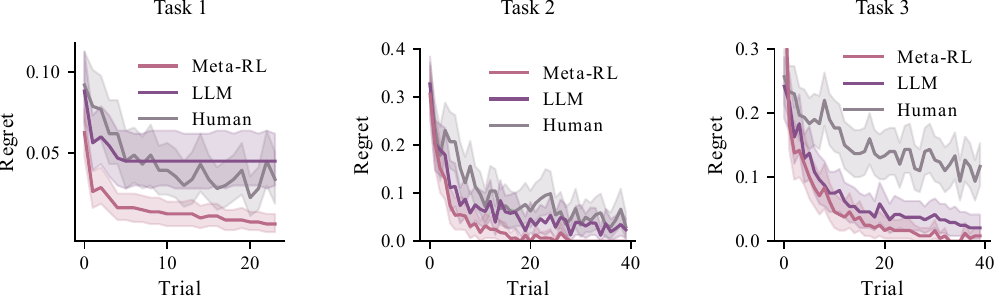}
    \caption{Average performance comparison of all three agents in terms of regret. In all tasks, the Meta-RL agent improves its performance over time fastest. Shaded areas correspond to 95\% CIs.}
    \label{fig:meta-rl-perf}
\end{figure}

After training to convergence, we tested the Meta-RL agent on the three experimental task under the same conditions as with the LLM. This idealized in-context learning agent outperformed the human as well as the LLM (see Figure~\ref{fig:meta-rl-perf}).

\section{Data and materials}
We will make all experimental data and code for this project available under the following link: [RETRACTED].

\end{document}